\documentclass{article}

\usepackage{microtype}
\usepackage{graphicx}
\usepackage{subfigure}
\usepackage{booktabs} 
\usepackage{caption}

\usepackage{hyperref}

\usepackage[accepted]{sysml2019}

\sysmltitlerunning{Collaborative Execution of Deep Neural Networks on Internet of Things Device}

\usepackage{amssymb}
\usepackage{pifont}
\newcommand{\cmark}{\ding{51}}
\newcommand{\xmark}{\ding{55}}

\newcommand*{\x}{\mathsf{x}\mskip1mu}

\usepackage{booktabs} 
\usepackage{array}
\usepackage{graphicx}
\usepackage{todonotes}
\usepackage{romannum}
\usepackage{multirow}

\usepackage{nicefrac}
\usepackage{mathtools}
\usepackage{bm}

\usepackage{amsmath}
\usepackage{algorithm}
\usepackage[noend]{algpseudocode}
\usepackage{txfonts}

\usepackage{enumitem}

\makeatletter
\renewcommand{\ALG@name}{Procedure}
\makeatother

\begin{document}

\twocolumn[
\sysmltitle{Collaborative Execution of Deep Neural Networks \\ on Internet of Things Devices}



\sysmlsetsymbol{equal}{*}

\begin{sysmlauthorlist}
\sysmlauthor{Ramyad Hadidi}{ga}
\sysmlauthor{Jiashen Cao}{ga}
\sysmlauthor{Micheal S. Ryoo}{goog}
\sysmlauthor{Hyesoon Kim}{ga}
\end{sysmlauthorlist}

\sysmlaffiliation{ga}{Georgia Institute of Technology, Atlanta, USA}
\sysmlaffiliation{goog}{Google}

\sysmlcorrespondingauthor{Ramyad Hadidi}{rhadidi@gatech.edu}

\sysmlkeywords{Machine Learning, SysML}

\vskip 0.3in

\begin{abstract}

With recent advancements in deep neural networks (DNNs), we are able to solve traditionally challenging problems. Since DNNs are compute intensive, consumers, to deploy a service, need to rely on expensive and scarce compute resources in the cloud. This approach, in addition to its dependability on high-quality network infrastructure and data centers, raises new privacy concerns. These challenges may limit DNN-based applications, so many researchers have tried optimize DNNs for local and in-edge execution. However, inadequate power and computing resources of edge devices along with small number of requests limits current optimizations applicability, such as batch processing. In this paper, we propose an approach that utilizes aggregated existing computing power of Internet of Things (IoT) devices surrounding an environment by creating a collaborative network. In this approach, IoT devices cooperate to conduct single-batch inferencing in real time. While exploiting several new model-parallelism methods and their distribution characteristics, our approach enhances the collaborative network by creating a balanced and distributed processing pipeline. We have illustrated our work using many Raspberry Pis with studying DNN models such as AlexNet, VGG16, Xception, and C3D.

\end{abstract}

]



\printAffiliationsAndNotice{} 

\section{Introduction and Motivation}

\label{sec:intro}

Deep neural networks (DNNs) are extending our capabilities to solve traditionally challenging problems such as computer vision, natural language processing, neural machine translation, and video recognition. While academia and companies are developing specialized hardware, these hardware are still expensive and they target high-performance computing (HPC) datacenters. Furthermore, conventional consumer-level devices, such as Internet of things (IoT) devices, lack the required performance to execute DNNs. As a result, to understand an environment, consumers need to offload these computation to cloud services. Such approach, in addition to a constant dependency on cloud services and high-quality network availability~\cite{gartner-iot-datacenter,lee:lee15,khan:khan12}, raises several new privacy concerns for users over their private data (e.g., 24/7 recordings of home security cameras)~\cite{IotSurvey}. At the same time, IoT devices are a perfect match for DNNs candidate applications (e.g., temperature sensors and smart cameras)~\cite{IotSurvey, gubbi2013internet}. This is because user's data remains and processed locally in the same network that is produced. How can we move the computations close to the edge by only using IoT devices, while providing an acceptable performance?

Our vision in this paper is to enable an efficient, local, and distributed computation of DNNs close to the edge by using IoT devices (i.e., resource-constrained devices). This is because a single IoT device cannot entirely handle the computations of DNNs. Although with some optimizations, such as weight pruning~\cite{yu:luk17,han:mao15} and precision reduction~\cite{cou:ben14, gon:li14, van:sen11}, we can run limited versions of the current models on IoT devices, with the advancement of DNNs and emergence of generalized model, the increase in demanded compute power for DNNs is not expected to stop. Therefore, exploring the distribution of DNN computation is essential. As discussed, since IoT devices are a great candidate for these DNN-based applications, this paper, by moving DNN computations closer to the edge, helps to address to achieve the following goals: (\romannum{1}) Reducing the dependability on cloud resources and high-quality network, (\romannum{2}) protecting consumer private data, (\romannum{3}) providing an alternative and cheaper solution to understand raw data locally, (\romannum{4}) developing a unified framework that is able to distribute any DNN model on \emph{existing} devices while providing real-time execution performance, (\romannum{5}) not being limited to a particular model or dataflow, and (\romannum{6}) decreasing the deployment time by providing general methods and avoiding model- and hardware-specific methods. 

We target IoT devices, the number of which have already outnumbered the world's population~\cite{gartner-iot, IotSurvey, sat17}, because these devices are generally idle for the most of the time. However, when performing the inference computations of DNNs using IoT devices, compared to the cloud, some important assumptions change. First, since the requests are local, we might not have enough data to process in parallel (i.e., no immediate data-level parallelism or batching). This means we cannot batch many requests immediately to amortize expensive costs of memory operations. Second, compared to HPC machines, IoT devices, besides having less computation power, have significantly smaller memories. Therefore, if the memory requirement of even a small computation task cannot fit in the memory of these devices, the execution performance suffers considerably. This is because, in such situations, the device uses off-chip storage as swap memory, which causes a huge slowdown.

In this paper, we propose a solution in which collaborative, low-power, and resource-constrained IoT devices perform distributed, real-time, and single-batch DNN-based recognition. By using these collaborative IoT devices, we generate and deploy a balanced data processing pipeline that is able to process DNN's computations efficiently. To address the challenge of limited memory space, we introduce several model-parallelism techniques for the common layers (convolution and fully-connected layers) of visual DNN models for reducing the memory footprint of their heavy computations. We discuss and analyze different DNN distribution methods. We also propose a heuristics to distribute DNNs on IoT devices by considering the memory requirement and amount of computation/communication to achieve the optimal performance. Moreover, we study prevalent visual DNN models such as image recognition (AlexNet~\cite{kri:sut12}, VGG16~\cite{sim:zis14-deep}, ResNet~\cite{he:zha16}, and Xception~\cite{chollet16}) and video recognition (C3D~\cite{du:bou15}). After examining various methods for model parallelism in fully-connected and convolution layers and their advantages and disadvantages, using our heuristics and monitoring tools, we create an evenly distributed data processing pipeline. For demonstration, we deploy our distributed system on an interconnected network of up to 11 Raspberry Pi 3s.

The reminder of this paper is organized as follows. In Section~\ref{sec:related}, we review prior work in this area. Section~\ref{sec:background} provides a background on convolution and fully-connected layers while introducing models used in the paper. Next, Section~\ref{sec:layers} explains model and data parallelism and gives a detailed exploration of model-parallelism methods for fully-connected and convolution layers. Then, in Section~\ref{sec:dist}, we discuss the processing pipeline and describe our heuristics in finding a near-optimal one. We evaluate our models in Section~\ref{sec:system}, and conclude the paper in Section~\ref{sec:conclusion} 

\section{Prior Work}

\label{sec:related}

Recently, with extensive large DNN models, distributing a single model has gained the attention of researchers~\cite{mao:chn17, tee:mcd17, had:cao18, kan:hau17, had:cao18:2, hadidi2018musical}. Large models need more memory, and when the memory requirement of a DNN model is larger than the system's memory, the performance of the model (both training and inference) suffers noticeably. More importantly, when executing DNNs on IoT devices versus companies datacenters, two important criteria changes: (\romannum{1}) first, a consumer, unlike large companies, cannot batch several requests and use more data parallelism. Therefore, all inferences are performed in a single batch mode which increases memory consumption per inference considerably. (\romannum{2}) Second, consumers does not have access to machines with high memory capacities, so more models suffers in performance. This is why recently some companies has released tools to alleviate this performance lost such as ELL library~\cite{ell} by Microsoft, and Tensorflow Lite~\cite{tensorflowLite} and MobileNets~\cite{mobilenet}. These libraries target devices such a Raspberry Pi, Arduino, and micro:bit. However, these tools are still in developments for single devices and do not distribute any computations on multiple devices. They aim for smaller number of weights, and convolution layers that have strides of two for reducing the dimensions of the input. Interestingly, some of tailored models in these implementation do not have any fully-conneted layers. Although such an effort might alleviate the overhead of DNNs on resource-constrained devices, the lower accuracy of the models in addition to the time to explore a specialized tailored model hinders the implementation of other models and increases the deployment time.

From the academic community, \cite{had:cao18} study, in which several robots collaborate together to perform distributed DNN computations, is the most similar to our work. The authors introduced only \emph{one method} that uses model parallelism on fully-connected layers, and use data parallelism for convolution layers. They do not study how the processing pipeline or model-parallelism methods for convolution layers helps the performance. Although they have provided an algorithm to distribute the tasks, we find that our heuristics with monitoring tools significantly shortens the time to find a near-optimal distribution. This is because their algorithm needs to have access to the entire profiled data, which takes a long time to gather and does not always covers all cases. By using the same technique in their work, we can also perform dynamic allocation during execution with similar concepts and tools. Another work, Neurosurgeon~\cite{kan:hau17}, dynamically partitions a DNN model between a \emph{single} edge device and the cloud, which incurs high network traffic and the increase the risk of privacy loss. Furthermore, the partitioning is always between the cloud and only \emph{one} edge device. DDNN~\cite{tee:mcd17} also tries to partition the model between edge devices and the cloud, but model retraining is necessary for each setting. In DDNN, Sensor devices (edge devices) only perform the first few layers in the network and the rest of the computation is offloaded to the cloud.

Another general direction is to reduce the overhead of DNNs by using methods such as weight pruning~\cite{yu:luk17,han:mao15,lin2017runtime}, resource partitioning~\cite{she:fer16, guo:yin17}, quantization and low-precision inference~\cite{cou:ben14, gon:li14, van:sen11,koster2017flexpoint}, and binarizing weights~\cite{li:zha16, cou:hub:16, rast:ord16}. Although these techniques reduce the overhead of DNNs, but require several additional steps that decrease the accuracy and enforce retraining the model. In fact, our work could be applied on top of these techniques to increase final performance as well. In other words, our work is orthogonal to these techniques in followings, which were not covered by previous studies. (\romannum{1}) We study resource-constrained devices with limited memory space, (\romannum{2}) we increase the real-time performance of single-batch DNN inferencing, (\romannum{3}) we introduce many methods for model parallelism, and (\romannum{4}) we design a collaborative system.

\section{Background}
\label{sec:background}
In this section, we provide an overview of convolution and dense layer computations to better understand how model and data parallelism in the next section applies to these layers. Note that we only introduce these two layers because they are among the most compute- and data-intensive layers~\cite{ven:ran17, rhu2016vdnn} in visual models. Then, we introduce our models that is used in the evaluation section.

\begin{figure}[h]
  \centering
  \vspace{-10pt}
  \includegraphics[width=0.55\linewidth]{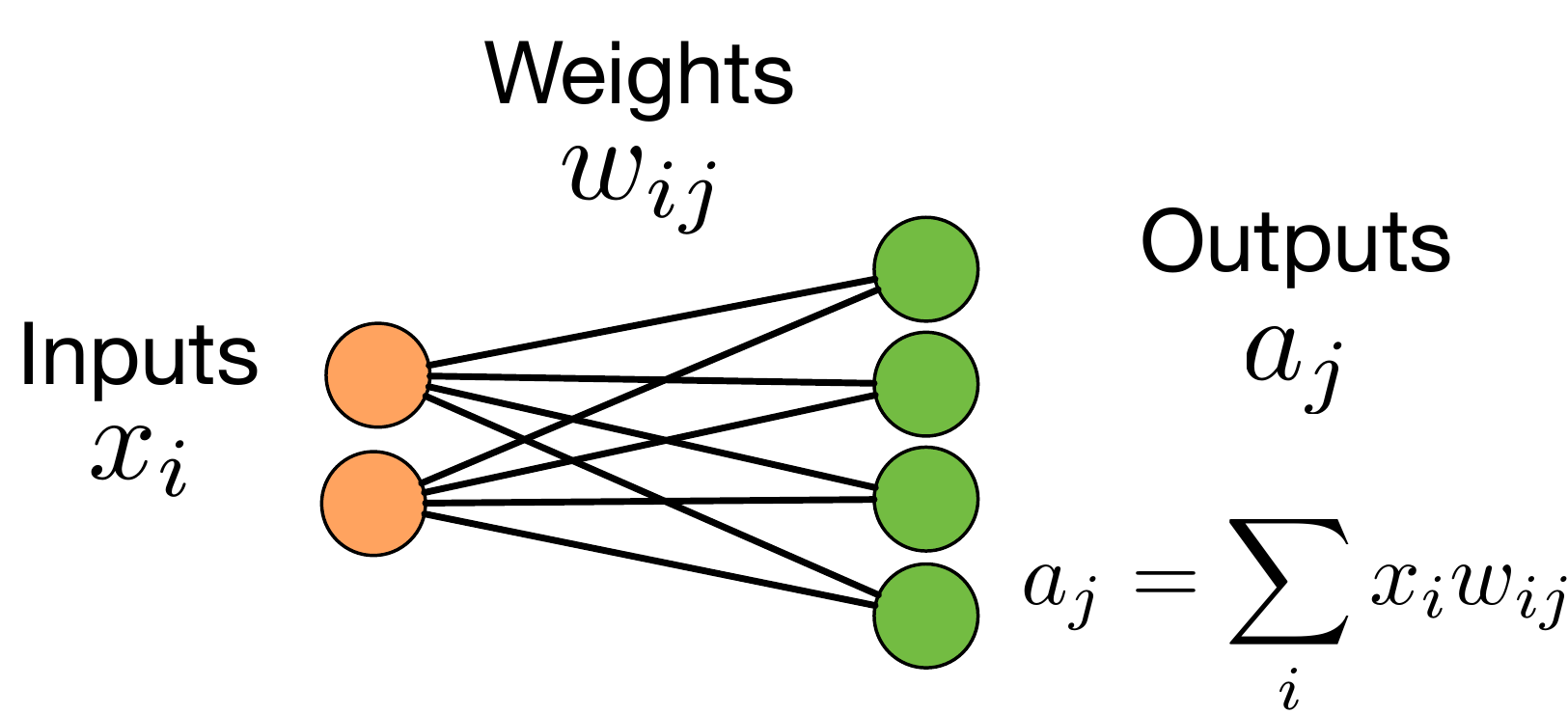}
  \vspace{-10pt}
  \caption{Dense layer computations.}
  \label{fig:dense-intro}
  \vspace{-5pt}
\end{figure}

\textbf{Dense Layer:}
In a dense or fully connected \texttt{(fc)} layer, the value of each output element is calculated from the weighted sum of all inputs. Figure~\ref{fig:dense-intro} depicts a dense layer with size four and input size of two. Each activation is calculated from the sum of inputs and weights products as
  $a_j= \sum_{i} x_i w_{ij} + b_j$,
in which $i$ is the input, $j$ is the output number, inputs are denoted as $x_i$, weights as $w_ij$, $b_j$ as biases, and $a_j$ as activations. This formula may be also written using matrix notations in $\bm{a = Wx+b}$. During training phase, the parameters of $\bm{W}$ and $\bm{b}$ are defined, and will be constant during inference phase.

%

\begin{figure}[h]
  \vspace{-5pt}
  \includegraphics[width=1.0\linewidth]{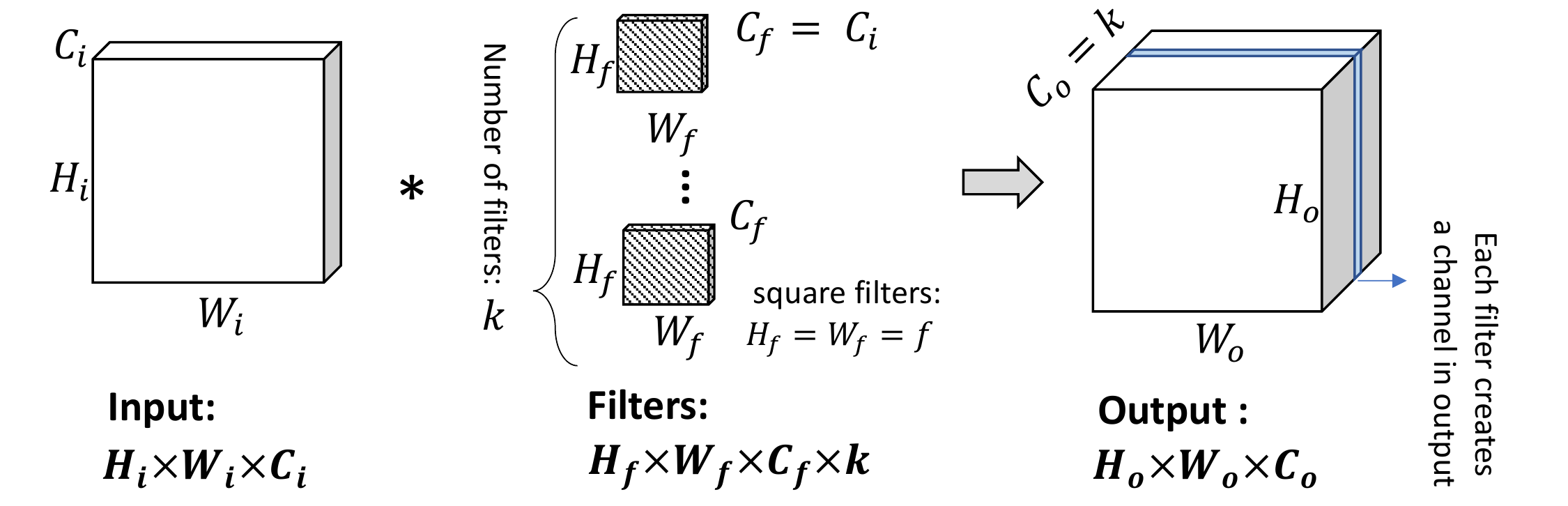}
  \vspace{-20pt}
  \caption{Convolution layer computations.}
  \label{fig:conv-filters}
  \vspace{-10pt}
\end{figure}

\begin{figure}[h]
  \includegraphics[width=1.0\linewidth]{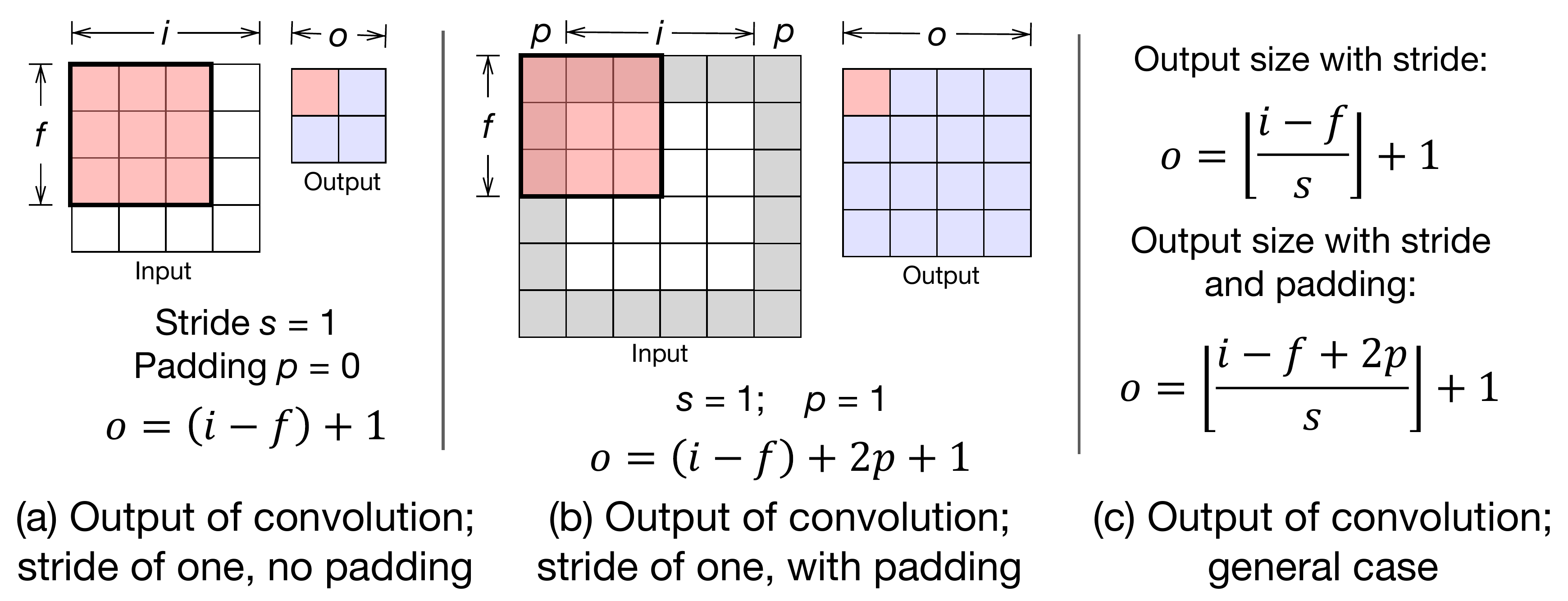}
  \vspace{-20pt}
  \caption{Convolution layer input and output sizes.}
  \label{fig:conv-output-size}
  \vspace{-8pt}  
\end{figure}

\textbf{Convolution Layer:}
DNN models are composed of several layers that process inputs. For visual models, usually, all the layers except last ones are convolution layers \texttt{(conv)}. A convolution layer applies a set of filters to a subset of inputs by sweeping each filter (i.e., kernel) over them. Each filter creates a channel, or depth (i.e., z-axis) of the output (Figure~\ref{fig:conv-filters}). The spatial dimensions (i.e., x- and y-axis) of the output is defined by four parameters: The size of input, filter, stride, and padding. In the simplest case of the convolution layer with just one filter, the filter slides across every position of the input producing one element per position. Figure~\ref{fig:conv-output-size}a illustrates the applying of a 3$\x$3 filter on an input of size 4$\x$4 with a unit stride ($s$), which is the shifting amount of the kernel, and zero padding ($p$), which is defined as extra zeros appended to the input to control the output size. Figure~\ref{fig:conv-output-size}b depicts the same example, but with a padding of one which means the the size of the new input is $i+2p$. Figure~\ref{fig:conv-output-size}c shows formulas to calculate the output size in general cases. In this paper, we use the same padding, which means the output size of a convolution layer is same as its input size. The same padding can be achieved by setting $p=\lfloor\nicefrac{f}{2}\rfloor$. In other cases, one can simply replace the output dimensions with the formulas in Figure~\ref{fig:conv-output-size}c. Note that Figure~\ref{fig:conv-output-size} shows 2D examples for convolution, similarly, we can extend such calculations to the 3D and 4D inputs and filters, usually used for time series.

%

\textbf{Other Layers:}
To introduce non-linearity, an activation layer ($\varphi$), such as Sigmoid or ReLU, is applied on the activations to create the input to the next layer, or $h_j = \varphi(a_j)$. This allows a model to learn complex functions. In addition, sometimes a pooling layer downsamples the input size and reduces the dimensions of data, such as max pooling \texttt{(maxpool)} or average pooling \texttt{(avgpool)} layers. These layers, compared to \texttt{fc} and  \texttt{conv} layers, are much less compute intensive, so we group them with their corresponding parent layer, which is the layer that produces their input.

\subsection{Models Overview}
We briefly overview model architectures we use in the paper. We specifically targeted tasks in computer vision because of their heavy computations and fast-paced advancements. Moreover, we tried our best to choose the prevalent and state-of-the-art models.

\begin{figure}[t]
  \vspace{-5pt}
  \includegraphics[width=1.0\linewidth]{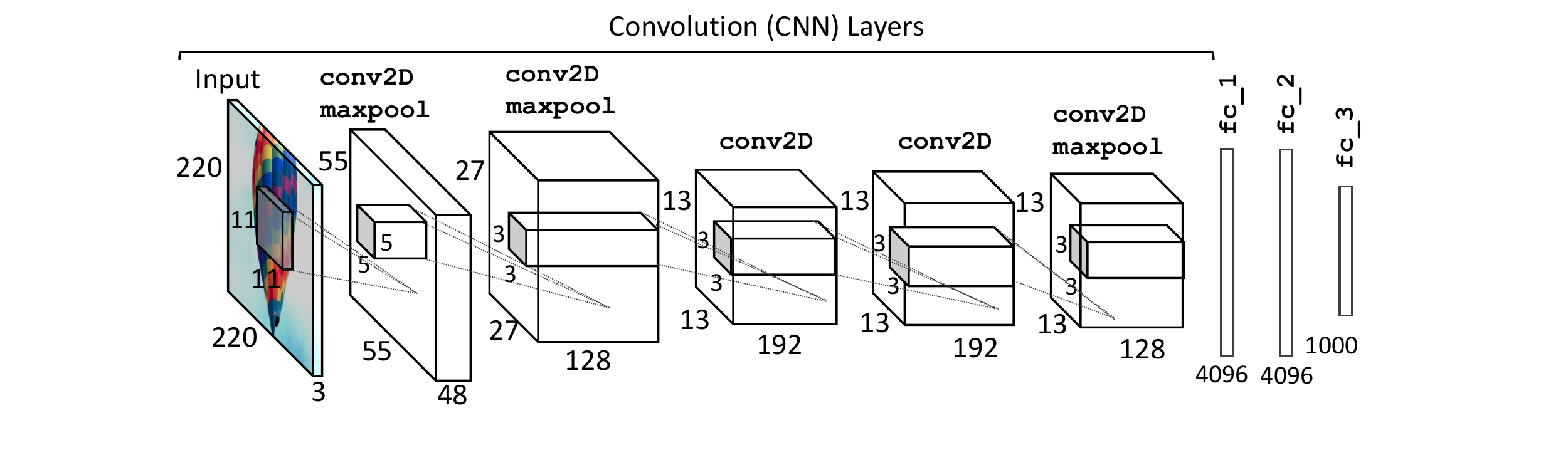}
  \vspace{-30pt}
  \caption{AlexNet.}
  \label{fig:alexnet}
\end{figure}

\begin{figure}[t]
  \vspace{-5pt}
  \includegraphics[width=1.0\linewidth]{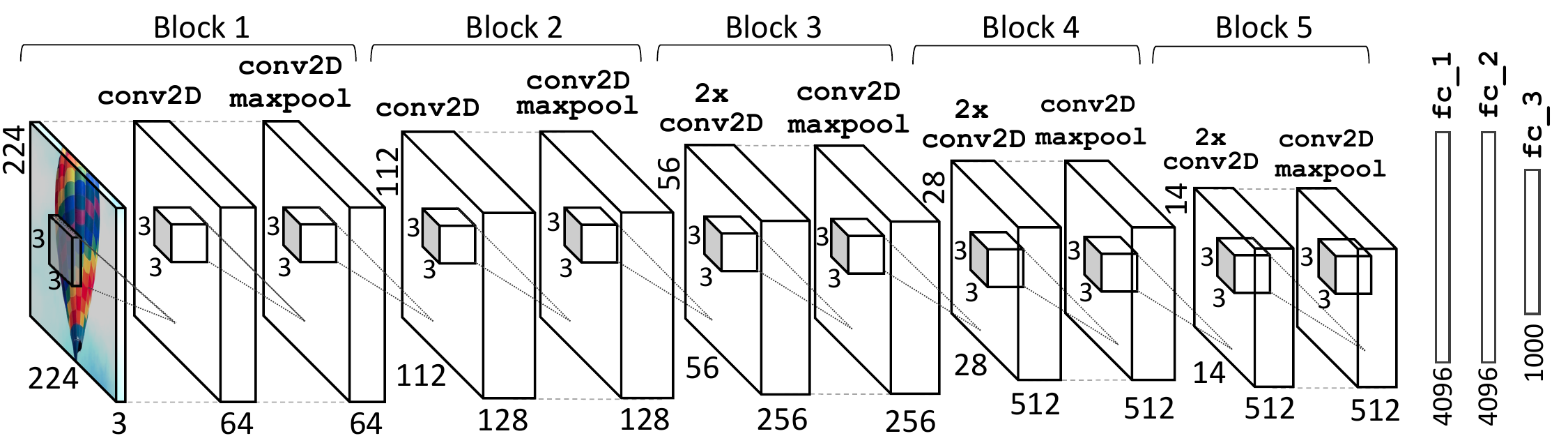}
  \vspace{-20pt}
  \caption{VGG16.}
  \label{fig:vgg16}
  \vspace{0pt}
\end{figure}

\textbf{AlexNet:} In 2012 ImageNet large-scale visual recognition challenge (ILSVRC), a challenge for image recognition task, AlexNet~\cite{kri:sut12} significantly outperformed all the prior competitors and won the challenge with a deeper CNN and more filters per layer. Figure~\ref{fig:alexnet} illustrates the model of the \emph{single} stream AlexNet, which consists of five convolution layers, and three dense layers. This model has a total of 40M parameters.
\textbf{VGG16:} Figure~\ref{fig:vgg16} depicts VGG16 model~\cite{sim:zis14-deep}, which has 16 layers, 13 convolution and three dense layers. As seen, VGG16 has a structured model, deeper convolution layers have more filters and smaller spatial dimensions. Number of parameters of VGG16 is around 140M, which is the highest in well-know image recognition models.

\begin{figure}[b]
  \vspace{-5pt}
  \includegraphics[width=1.0\linewidth]{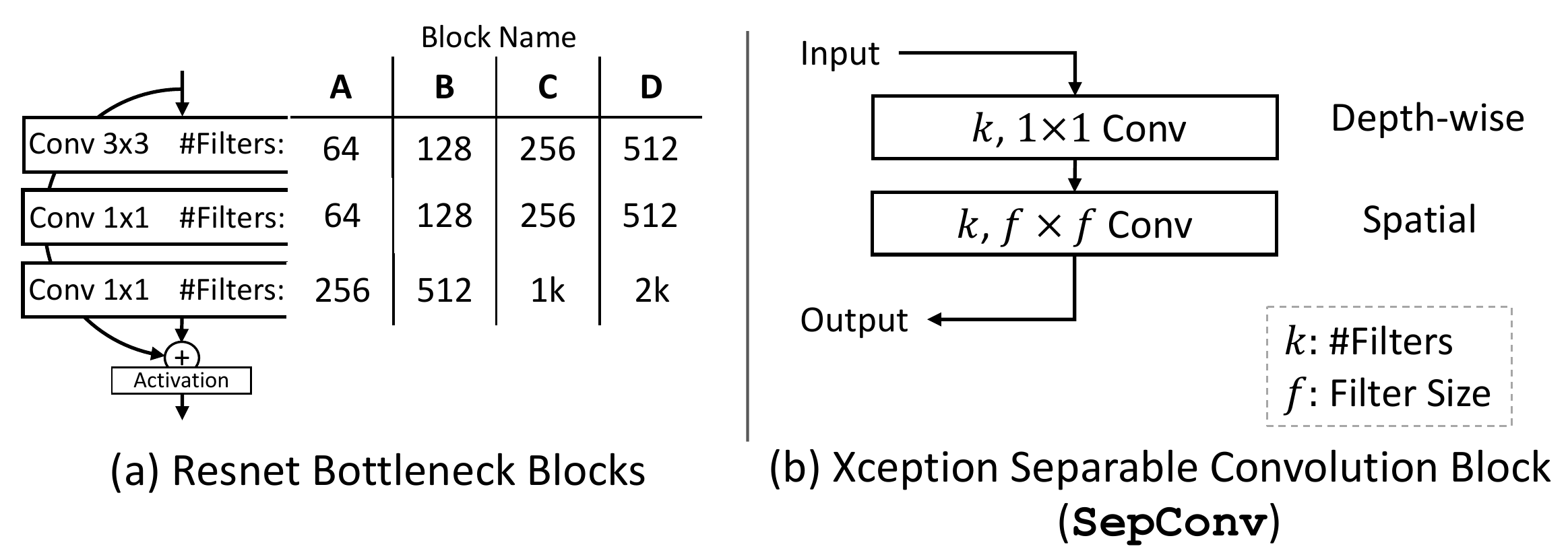}
  \vspace{-20pt}
  \caption{Building blocks of Resnet50 and Xception.}
  \label{fig:blocks}
  \vspace{0pt}
\end{figure}

\textbf{ResNet:} Residual neural network (ResNet)~\cite{he:zha16} introduced ``skip-connection'' for training deeper network in 2016. In this paper, we used ResNet50 that has 50 layers. Figure~\ref{fig:blocks}a illustrates basic blocks for ResNet. This model is residual in a sense that a shortcut connection skips a block, and makes training easier for such a deep model. Although ResNet50 is a deep model with several layers, the total number of parameters are 25M. 

\begin{figure}
  \vspace{-5pt}
  \includegraphics[width=1.0\linewidth]{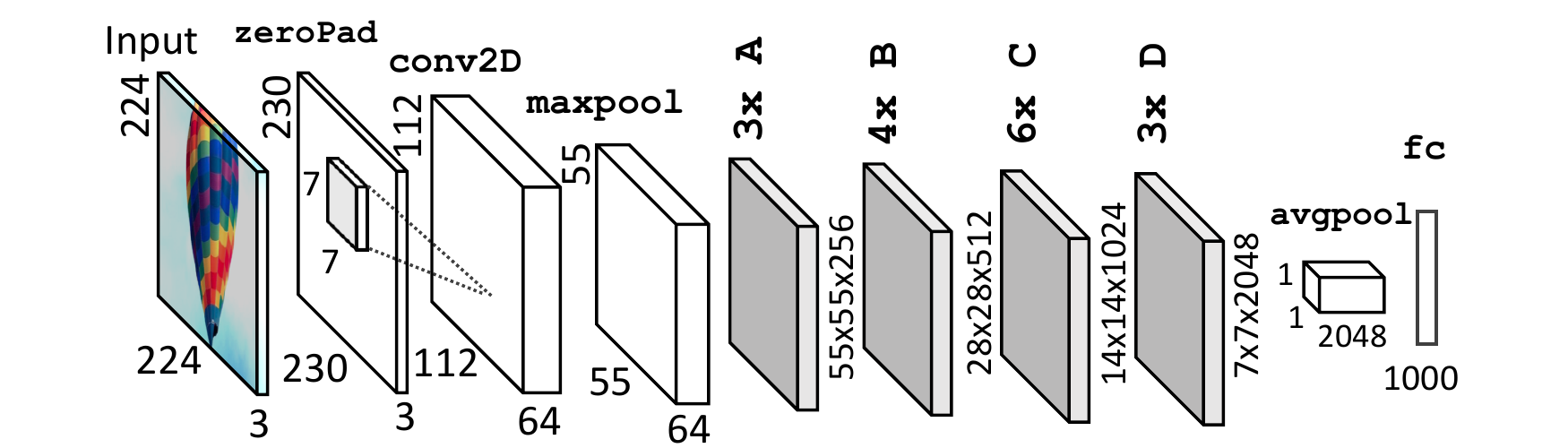}
  \vspace{-20pt}
  \caption{Resnet50.}
  \label{fig:res50}
  \vspace{-5pt}
\end{figure}
 
\begin{figure}[t]
  \vspace{0pt}
  \includegraphics[width=1.0\linewidth]{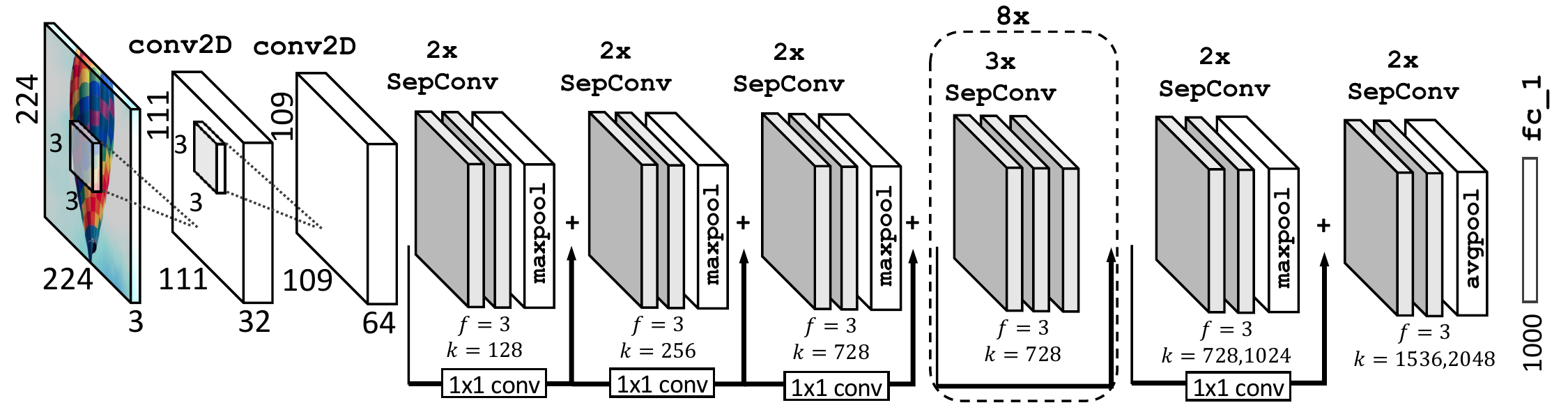}
  \vspace{-20pt}
  \caption{Xception.}
  \label{fig:xception}
  \vspace{-5pt}
\end{figure}

\textbf{Xception:} The most recent and accurate image recognition model among the models we used is Xception~\cite{chollet16}. This model is based on Inception V3~\cite{sze:van16}. Xception extends Inception module with a vision to process cross-channel and spatial correlations independently. Therefore, Xception introduces a special convolution layer, shown in Figure~\ref{fig:blocks}b, separable convolution~\cite{chollet16}, that its mapping of cross-channel and spatial correlations is decoupled. Separable convolution first performs cross-channel (i.e., depth-wise) convolution over input channels, and then performs an independent spatial convolution on each of the outputs. Figure~\ref{fig:xception} shows Xception model, with 34 separable convolution layers. The total number of parameters for this model is 23M.
  
\begin{figure}[b]
\vspace{0pt}
  \includegraphics[width=1.0\linewidth]{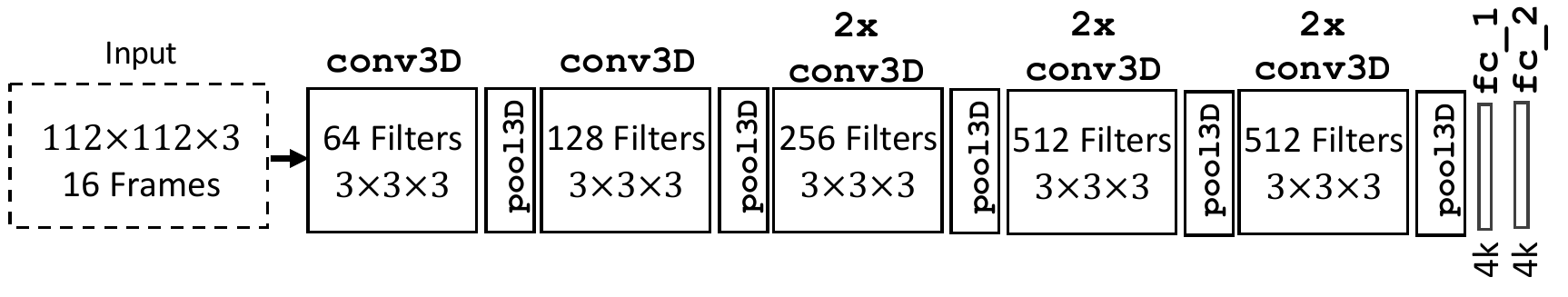}
  \vspace{-25pt}
  \caption{C3D.}
  \label{fig:c3d}
  \vspace{-5pt}
\end{figure}

\textbf{C3D:} Convolution 3D (C3D)~\cite{du:bou15} model is designed to process videos and has been used in action recognition and scene classification tasks. To learn spatio-temporal features, C3D model uses 3D convolutions, which produce an output volume instead of a 2D output per filter. Compared to conventional convolution layer, an additional sweep along the z-axis  creates a volume in the output. Figure~\ref{fig:c3d} shows C3D model, which consists of eight 3D convolution layers. The total number of parameters for this model is 80M.

\section{Distributing and Parallelizing Inference}
\label{sec:layers}

%
%
%
%
%
%
%
%
%
%
%
%
%

%
%
%
\renewcommand{\arraystretch}{0.9}
\begin{table*}[h]
\vspace{-0pt}
\centering
\small
\caption{Characteristics of model parallelism methods for dense layers for a layer of input dimension $d_i$ and output dimension $d_o$.}
\vspace{2pt}
\begin{tabular}{c || c | c | c | c | c | c | c}
  \toprule
   \multirow{2}{*}{Name} 
      & \#Node
      & Distributed 
      & Multipication
      & Reduction
      & Weights
      & Communication
      & Merge\\
   
      & 
      & Activation
      & {\scriptsize (per node)}
      & {\scriptsize (per node)}
      & {\scriptsize (per node)}
      & {\scriptsize (total-per inference)}
      & Operation \\
  \midrule
  
   No Splitting
      & 1
      & {\scriptsize N/A}
      & $d_i d_o$
      & $d_o$
      & $d_i d_o$
      & $d_i + d_o$
      & {\scriptsize N/A}\\
  \midrule
      
  Output Splitting
      & $n$
      & \cmark
      & $\frac{d_o}{n} d_i$
      & $\nicefrac{d_o}{n}$
      & $n d_i$
      & $nd_i + d_o$
      & {\small Concat}\\   
  \midrule
  
  Input Splitting
      & $n$
      & \xmark
      & $\frac{d_i}{n} d_o$
      & $d_o[|(\nicefrac{d_i}{n}-1)]$
      & $n d_o$ 
      & $d_i + nd_o$
      & {\small Sum}\\
  \bottomrule
\end{tabular}
\label{table:dense-methods}
\vspace{-8pt}
\end{table*}
\renewcommand{\arraystretch}{1}
In this section, we overview our methods for distributing and
parallelizing inference computations for dense and convolution
layers. We examine two general directions: data parallelism and model
parallelism. In data parallelism, we rely on the presence of many data
inputs to distribute/parallelize the computations. This direction
enables us to increase the number of inferences per second while
maintaining a constant time to process each input. In model
parallelism, which is applicable to the computations required for a
single input, the computations is distributed/parallelized over
multiple compute nodes. By following this direction, we reduce the
time to process an input. 

Since the DNNs have multiple of layers, it has a built-in pipeline
parallelism. Hence, the first step is to divide a model into multiple
devices by layers (or a group of layers) to utilize the pipeline
parallelism.  These layers process the input sequentially and the
output of each layer is dependent on the output of its previous
layer(s). Thus, we must correctly maintain this dependency between
layers. Using the pipeline parallelism, we can increase the
throughput of computation while the latency for each computation
remains the same. We improve the performance further by applying data-level and
model-level parallelism on top of pipeline parallelism. 

\textbf{Going Further than Data Parallelism:}
Data parallelism is already introduced by \cite{had:cao18} for dense
and convolution layers for real-world models. But, only applying data
parallelism would not always work for resource-constrained devices and
in the edge device scenarios. Data parallelism duplicates a node that
performs the same computation. Since the computation is the same, but
on a different input data, memory footprint is not reduced. In fact,
this is one of the main reasons why data parallelism is not enough for
distributing and parallelizing DNN computations. This is because:
(\romannum{1}) For sufficiently large layers, just the duplication of
devices would not give us a good performance benefit because the
entire data is not loaded to the memory, and a device pays a high cost
for accessing the off-chip storage (i.e., swap). (\romannum{2}) Data
parallelism needs a stream of input data. But, in some cases such a
single image inference or sentence translation, we only have one data
input or input injection frequency is low. (\romannum{3}) To create a
balanced and efficient data processing pipeline in our distributed
system, we need a balanced pipeline design (i.e., the amount of
computation per each node should be almost the same. ) However, data parallelism is not flexible in adjusting the amount of computation on each node.

\textbf{Model Parallelism:}
Model parallelism is based on this fact that since the computations of each given layer to calculate its output are independent from each other, we can parallelize the computations. For instance, in a convolutional layer, the computation of each element in the output is independent from all other elements. Therefore, in model parallelism, we can exploit such intra-layer independency of computations to increase parallelism. In fact, employing such deeper level parallelism, compared to data parallelism, needs a knowledge of how each layer does its computations, and how parallelism affects data communication, computations, and aggregation. In summary, in a DNN model, model parallelism is to distribute/parallelize the computations of a single input. In the following, we introduce our model parallelism methods for dense and convolution layers.

\subsection{Model Parallelism for Dense Layers}
\begin{figure}[h]
  \includegraphics[width=1.0\linewidth]{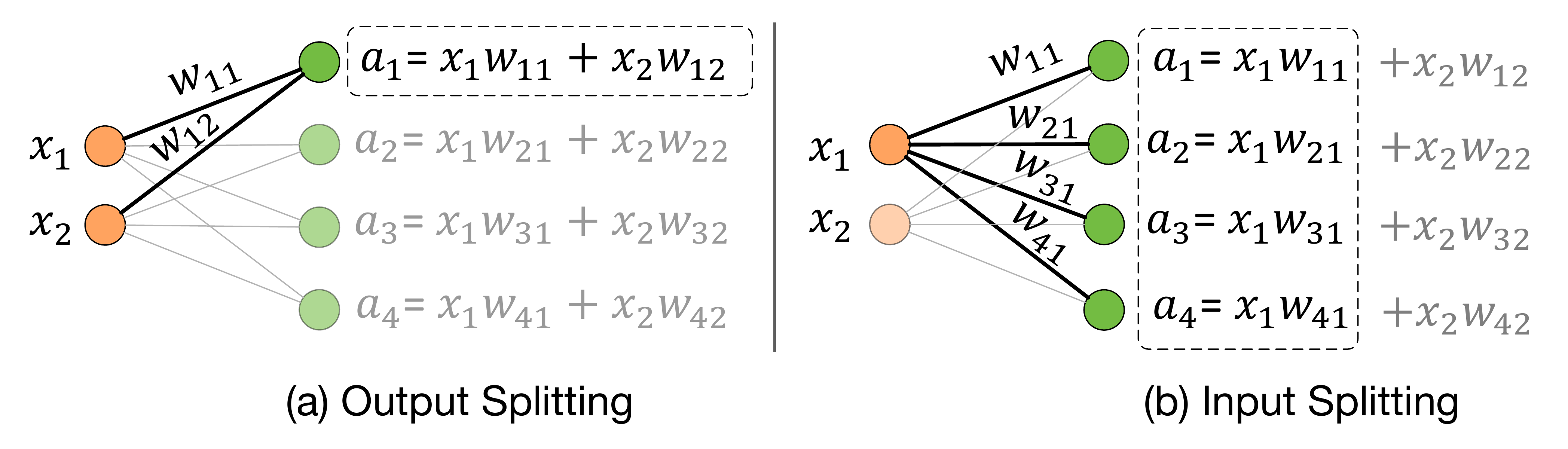}
  \vspace{-20pt}
  \caption{Model parallelism methods for dense layers.}
  \label{fig:fc-model-parallelism}
  \vspace{-0pt}
\end{figure}

In a dense layer, since the computations of each activation ($a_j$) is independent from other activations, we can parallelize the computations of a dense layer. We describe two model parallelism methods specific to dense layers: Output and input splitting, shown in Figure~\ref{fig:fc-model-parallelism}a and b, respectively. In output splitting, we parallelize the computation of each activation, while we transmit all input data to all devices. Figure~\ref{fig:fc-model-parallelism}a highlights a node and its computations to derive its activation. As seen, for each node, we need to transmit all the inputs. Moreover, each node holds the weights corresponding to its activations. Later, when each node is done with its computations, we merge the results. The merge consists of concatenating values in a correct order. In addition, we can apply activation function (e.g., RELU, Sigmoid) either on each node, or after the merging.
 In input splitting, a node computes a partial part of all activations. Figure~\ref{fig:fc-model-parallelism}b illustrates an example in which a node computes the half of required multiplications for all the activations. In this method, we transmit a part of the input to each node. Furthermore, each node holds the weights corresponding to the input split that processes. Later, when each node is done with its computations, we merge the results by adding all of the corresponding partial sums. Thus, the merge consists of a reduction operation (i.e., summation). However, contrary to the output splitting method, we cannot apply activation function before the merge. Mentioned methods may also be mixed which creates a spectrum of methods, however, in this paper we focus on extreme cases of this spectrum.

\begin{figure}[h]
  \includegraphics[width=1.0\linewidth]{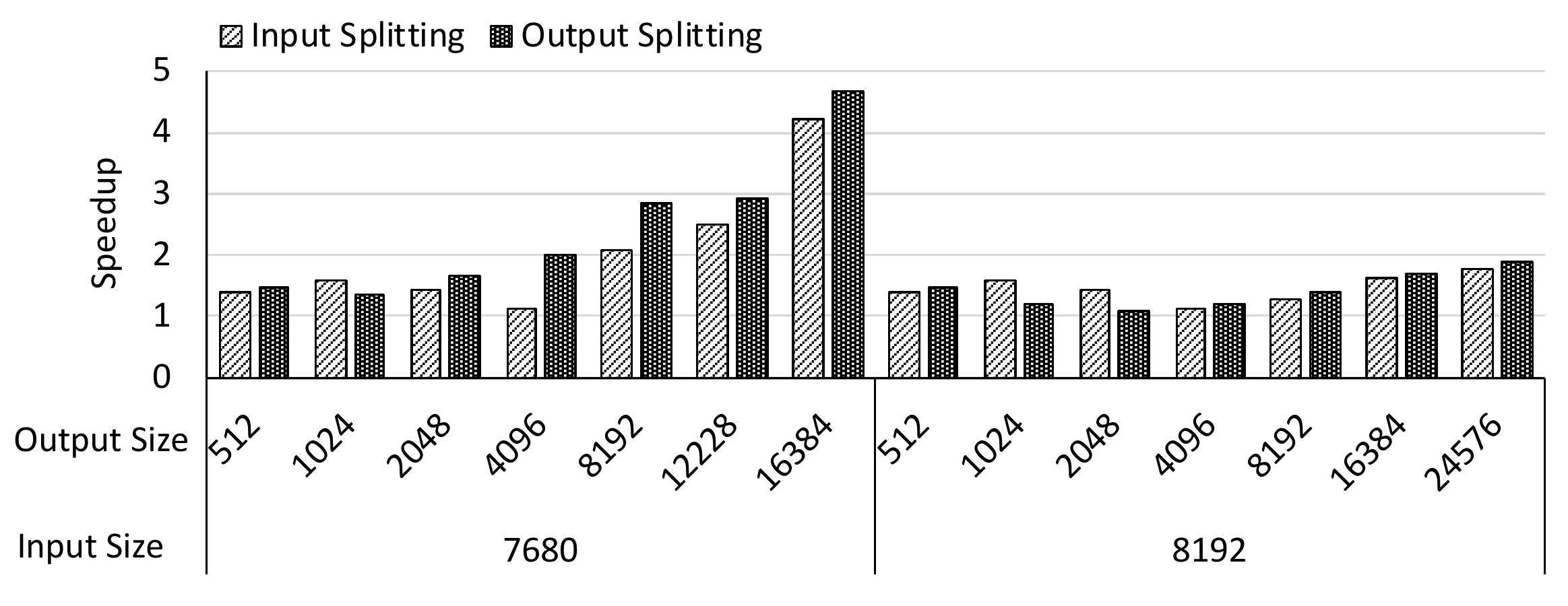}
  \vspace{-20pt}
  \caption{Performance of model-parallelism methods on two devices for different dense layers.}
  \label{fig:fc-methods}
  \vspace{-0pt}
\end{figure}

 A more detailed summary of the mentioned methods are presented in
 Table~\ref{table:dense-methods}. These methods trade communication
 with the memory footprint. This is because each node holds a part of
 the weights, but need to transmit more variables. The more detailed
 examination is done in the table where $n$ is the number of the
 nodes, and $d_i$ and $d_o$ are input and output dimensions,
 respectively. As seen, both methods divide the memory footprint
 (i.e. saved weights) and the amount of multiplications. Input
 splitting slightly increases the number of reductions because
 computing the partial sums are necessary on each node if the node
 receives more than one input element. Furthermore, output and input
 splitting methods have a communication overhead of $(n-1)d_i$ and
 $(n-1)d_o$, respectively. We run a series of dense layers in
 Figure~\ref{fig:fc-methods} on a single device, and their distributed
 versions on two devices (in total four devices, with initial sender
 and final receiver). We cover a range of 512 to 16384 in output
 sizes, and two input sizes, 7680 (not power of two) and 8192 (power
 of two). As seen, for the input size of 7680 and large output sizes,
 we achieve super-linear speedups. This is because in these cases,
 slow off-chip storage (i.e., swap) is used. On the other hand, for
 input size of 8192, the baseline DNN framework can optimize accesses
 and avoid swap activities by tiling. The baselind DNN framework
 optimizes the swap space accesses, it cannot always hide the cost
 such as 7680 as we shown or a larger size than 8192. 
Thus, model parallelism helps us in avoiding such costs. Furthermore, other speedup values in the figure is less than the ideal value of two because each distribution has a communication cost. Input splitting has mostly lower performance than output splitting since it cannot apply activations locally.

 %

\renewcommand{\arraystretch}{0.9}
\begin{table*}[h]
\centering
\footnotesize
\vspace{-0pt}
\caption{Model parallelism methods for convolution layers (assuming same padding).}
\vspace{5pt}
\begin{tabular}{ c|| c| c| c| c| c| c| c| c| c }
\toprule  

    \multirow{2}{*}{Name}
    & \footnotesize{Division}
    & \multirow{2}{*}{\footnotesize{\#Nodes}}
    & \footnotesize{Distributed}
    & \footnotesize{Weights}
    & \footnotesize{Input}
    & \footnotesize{Filters}
    & \footnotesize{Output}
    & \footnotesize{Communication}
    & \footnotesize{Merge}
    \\
    
    ~
    & \footnotesize{Factor}
    &
    & \footnotesize{Activation}
    & \scriptsize{(per node)}
    & \scriptsize{(per node)}
    & \scriptsize{(per node)}
    & \scriptsize{(per node)}
    & {\scriptsize (total-per inference)}
    & \footnotesize{Operation}
    \\
    \midrule

    \small{Baseline}
    & {\scriptsize N/A}
    & 1
    & {\scriptsize N/A}
    & \footnotesize{$k C_i f^2$}
    & \footnotesize{$H_i W_i C_i$}
    & $k C_i f^2$
    & \footnotesize{$H_i W_i k$}
    & $(C_i + k)(H_i W_i)$
    & {\scriptsize N/A}
    \\
    \midrule

    \small{Channel}
    & $k'$ $\frac{\text{filters}}{\text{node}}$
    & $\lceil \frac{k}{k'} \rceil$
    & \cmark
    & \footnotesize{$k' C_i f^2$}
    & \footnotesize{$H_i W_i C_i$}
    & $k' C_i f^2$
    & $H_i W_i k'$
    & $(\lceil \frac{k}{k'} \rceil C_i + k)(H_i W_i)$
    & \footnotesize{Concat}
    \\
    \midrule

    \small{Spatial}
    & $d$ $\frac{\text{part}}{\text{dimension}}$
    & $d^2$
    & \cmark
    & \footnotesize{$k C_i f^2$}
    & \footnotesize{Eq.\ref{equ:spatial_input}}
    & $k C_i f^2$
    & $\frac{1}{d^2} H_i W_i k$
    & $(d^2 Eq.\ref{equ:spatial_input} + k)(H_i W_i)$
    & \footnotesize{Concat}
    \\
    \midrule

    \small{Filter}
    & $C_b$ \scriptsize{batches}
    & $\lceil \frac{C_i}{C_b} \rceil$
    & \xmark
    & \footnotesize{$k C_b f^2$}
    & \footnotesize{$H_i W_i C_b$}
    & $k C_b f^2$
    & $H_i W_i k $
    & $(C_i + k\lceil \frac{C_i}{C_b} \rceil)(H_i W_i)$
    & \footnotesize{Sum} \\
    \bottomrule
    
\end{tabular}
\vspace{-0pt}
\label{table:conv-methods}
\end{table*}
\renewcommand{\arraystretch}{1}
\renewcommand{\arraystretch}{0.5}
\begin{table*}[h]
\centering
\footnotesize
\vspace{-5pt}
\caption{Comparisons of model parallelism methods for convolution layers}
\vspace{5pt}
\begin{tabular}{ >{\centering\arraybackslash}p{1.2cm} |
                 >{\centering\arraybackslash}p{4.8cm} | 
                 >{\centering\arraybackslash}p{4.8cm} | 
                 >{\centering\arraybackslash}p{4.8cm} }

    & Channel Splitting
    & Spatial Splitting
    & Filter Splitting 
    \\
    \toprule
    
    Input
    & Entire input is copied
    & Input is divided spatially
    & Input is divided channel-wise 
    \\
    \midrule
    
    Filters
    & Some filters are saved
    & All filters are saved
    & Part of all filters are saved
    \\
    \midrule
    
    Output
    & Each node calculates a channel
    & Each node calculates a spatial region
    & Each node calculates a partial output
    \\
    \midrule
    
    Overhead
    & Input is copied across all nodes
    & Input overlapping elements
    & Output partial sums
    \\
    \bottomrule
    
\end{tabular}
\vspace{-8pt}
\label{table:conv-methods-ovreall}
\end{table*}
\renewcommand{\arraystretch}{1}


\subsection{Model Parallelism for Convolutional Layers}

As discussed, in a convolution layer, each filter creates a channel in the output data. As Figure~\ref{fig:conv-filters} illustrates, let us assume the dimensions of input, filters, and output is as $H_i \x W_i \x C_i$, $H_f \x W_f \x C_f \x k$, and $H_o \x W_o \x C_o$, respectively. The depth of filters is defined by the depth of input, or $C_f = C_i$. Here, without loss of generality, we assume square filters, $H_f = W_f = f$. The number of channels in output is defined by number of filters, or $C_o = k$. Each filter contains $C_i f^2$ weights that are set during training. \emph{Per output element}, each filter performs $C_i f^2$ multiplications of its weights and input values, and one reduction operation. So, for $k$ filters in a convolution layer, per output element, we perform $k C_i f^2$ multiplications and $k$ reductions. Therefore, from Figure~\ref{fig:conv-output-size}, total number of multiplications and reductions in a convolution layer for \emph{all elements} are as below:
\begin{equation}
\vspace{0pt}
\begin{split}
\text{\footnotesize{Multiplications:}}~ H_o W_o k C_i f^2 &\xRightarrow[]{\text{\tiny{Same Padding}}} H_i W_i k C_i f^2 \\
\text{\footnotesize{Reductions:}}~ H_o W_o k &\xRightarrow[]{\text{\tiny{Same Padding}}} H_i W_i k.
\end{split}
\vspace{0pt}
\end{equation}
For a single inference, the amount of communication is the sum of number of input and output elements, or $(H_i W_i C_i) + (H_i W_i k) = (C_i + k)(H_i W_i)$.
In the reset of this section, we describe our specific methods of model-parallelism for convolution layers. Since each method has its own advantages and disadvantages depending on the target convolution layer, choosing the best method requires careful considerations. To summary, Table~\ref{table:conv-methods} provides a detailed overview of discussions in this section.

\begin{figure}[t]
  \vspace{-0pt}
  \includegraphics[width=1.0\linewidth]{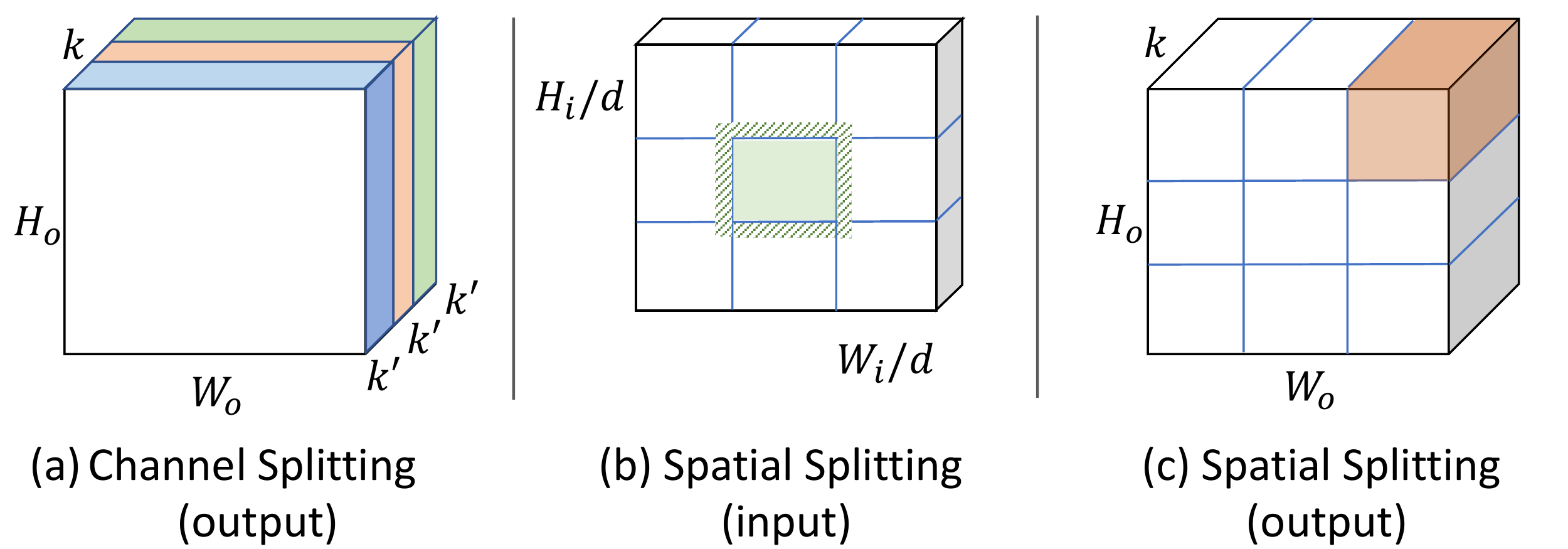}
  \vspace{-20pt}
  \caption{Convolution layer channel and spatial splitting.}
  \vspace{-0pt}
  \label{fig:conv-model-parallelism1}
\end{figure}

\textbf{Channel Splitting:}
In channel splitting, each node calculates a non-overlapping set of
channels in the output. In other words, each node only processes,
namely $k'$ filters, $k' \leq
k$. Figure~\ref{fig:conv-model-parallelism1}a shows an example output
of this method with three nodes. Since $k'$ filter is processed per
node, we need a total of $\lceil \nicefrac{k}{k'} \rceil$ nodes. Each
node only needs its set of $k'$ filters, but because each node takes
the whole input, each needs a copy of the input data. Filters are
divided, so each node saves the weights of its dedicated filters, or
$k' C_i f^2$. The total number of multiplications and reductions
remains the same, and each node handles ${\lceil \nicefrac{k}{k'}
  \rceil}^{-1}$ part. At the end, when every node is done with its
computations, we concatenate their data depth-wise which is in
$O(k)$. For the output, the total number of output elements to be
transferred is $H_i W_i k$. We have the option to apply activation
function on each node or after the merging, since activation function
applies on every element independently. In total, based on Table~\ref{table:conv-methods}, we pay $(\lceil \nicefrac{k}{k'} \rceil C_i - 1) H_i W_i$ in communication overhead, since we need to transmit a copy of the input to all nodes.

\textbf{Spatial Splitting:}
In spatial splitting, instead of splitting the filters, we split the input spatially (in x- and y-axis). Let us assume that we split each dimension in $d$ parts, so we have a total of $d^2$ parts\footnotemark, as shown in Figure~\ref{fig:conv-model-parallelism1}b. We transmit each part of the input to a node. Furthermore, we need to extend each region for $\lfloor \nicefrac{f}{2} \rfloor$ more overlapping elements with neighboring parts, so that we can do convolution on the borders. Therefore, the number of input data elements to be transmitted per node is:
\begin{equation}
  \lceil \frac{1}{d^2} \rceil H_i W_i C_i + 4\lfloor \nicefrac{f}{2} \rfloor(d^2-d),
  \label{equ:spatial_input}
\end{equation}
in which the first term represents the splitted input, and the second
term represents the numbers of extra overlapping elements. Compared to
channel splitting in which we transmit of copy of input to all nodes,
here we pay the extra overhead for only the overlapping parts. Since
each node processes all filters, each needs a copy of all
weights. Hence, the total number of filter elements to be transmitted
is $d^2 k C_i f^2$. However, note that this is a one-time cost for all
inferences. The total number of multiplications and reductions is the
same in total, and each node process only $\nicefrac{1}{d^2}$
part. When the computation of each node is done, we concatenate their
output spatially. The concatenation is in order of the total number of
parts. Similar to previous method, the total number of output elements
to be transferred is $H_i W_i k$. We have the option to apply
activation function either on each node or after the merging. As
discussed, the communication overhead for spatial splitting is only
for overlapping parts, which approximately is $4d^2\lfloor
\nicefrac{f}{2} \rfloor(d^2-d)$. Since usually filter size is small,
this overhead is not significant. Spatial splitting has another
advantage, which is to generate a part of output, we do not need to
merge all the results. Therefore, in constructing a parallelized model
while maintaining correctness, we can process a few convolution layers sequentially without merging their result back after every one layer.

\footnotetext{Here, for simplicity we divide each dimension to equal parts, which only allows square numbers for the number of the nodes. In our implementations, we implement the general version with unequal parts for which any number of nodes is possible. }

\begin{figure}[h]
  \vspace{-0pt}
  \includegraphics[width=1.0\linewidth]{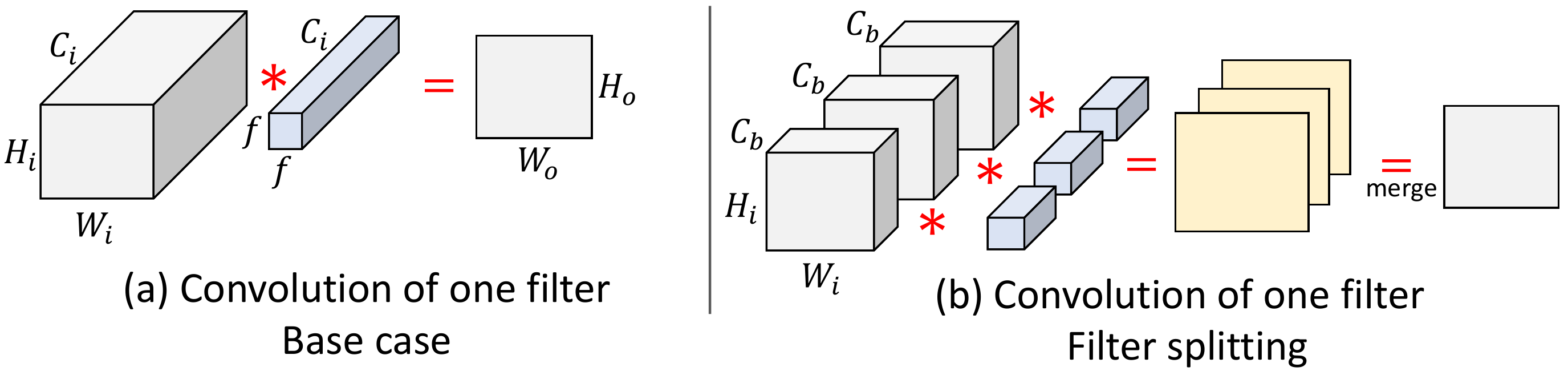}
  \vspace{-20pt}
  \caption{Convolution layer filter-splitting method.}
  \vspace{-5pt}
  \label{fig:conv-model-parallelism2}
\end{figure}

\begin{figure*}
  \vspace{-0pt}
  \includegraphics[width=1.0\linewidth]{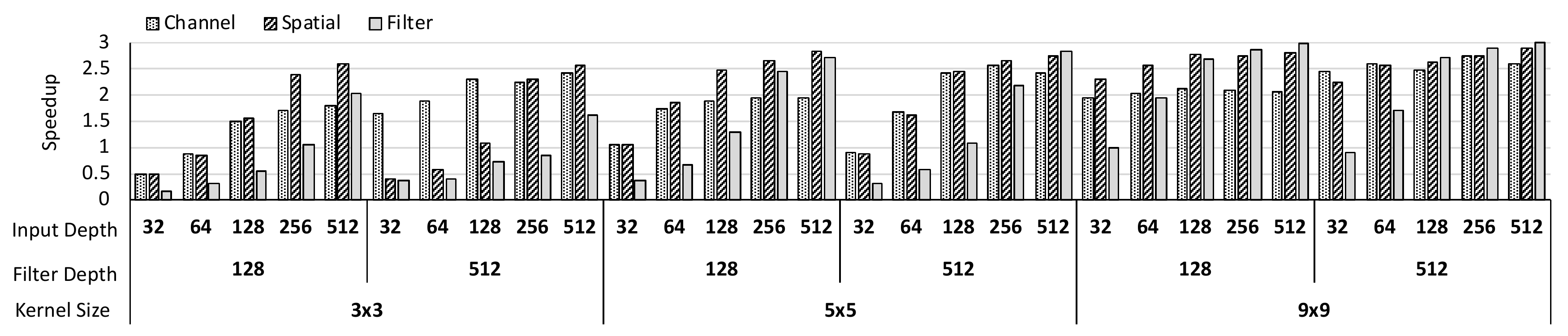}
  \vspace{-20pt}
  \caption{Performance of model-parallelism methods on three devices for different convolution layers.}
  \vspace{-8pt}
  \label{fig:conv-methods-all}
\end{figure*}

\textbf{Filter Splitting:}
In filter splitting, both input and filter are splitted channel-wise
in batches of size $C_b$. Figure~\ref{fig:conv-model-parallelism2}a
illustrates the base case in the convolution of one filter which
produces a single channel in the
output. Figure~\ref{fig:conv-model-parallelism2}b illustrates same
filter in the filter splitting method. Both the input and filter is
divided to three parts, each of which is processed separately. Since
there is a one-to-one correspondence between input and filter
elements, each node computes a partial output. In the end, to create
the final output, we need to sum all corresponding elements and apply
the activation function. If we denote input channel size as $C_i$, then we need a total of  $\lceil \nicefrac{C_i}{C_b} \rceil$ nodes. Since input is splitted channel wise, total number of input elements transfers is without an overhead, or $H_i W_i C_i$ in total. Similarly, each node only saves its dedicated channels of all filters, so memory footprint is also divided. But, since each node sends a partial output to the merging node, there is overhead of $(k\lceil \frac{C_i}{C_b} \rceil -1)(H_i W_i)$ for transmitting output elements compared to the baseline. In addition, to create the final output, we need to perform $k\lceil \frac{C_i}{C_b} \rceil$ reductions. The concatenation is in $O(\nicefrac{C_i}{C_b})$.

\textbf{Methods Comparison:}
Now that we know different methods for model parallelism, how should
we choose the best performing one for a specific convolution layer? A
better comparison of these methods is presented in
Table~\ref{table:conv-methods-ovreall}. For instance, channel
splitting has an overhead of copying the input, whereas filter
splitting has to transmit partial sums. The impact strength of these
differences on the performance is defined by the properties of a
convolution layer. As illustration, in
Figure~\ref{fig:conv-methods-all}, we run a convolution layer with the
kernels $3\x3, 5\x5, 9\x9$, filter depths 128 and
512, and various input depths with $128\x128$ inputs. We distributed the layer on three
Raspberry Pis using the mentioned methods (in total five devices, with
initial sender and final receiver). Speedups are relative to single 
deivce execution. We see that in the kernel
$3\x3$ and filter depth 128, smaller input depths have no
speedup. This is because the amount of computation per node after
distribution is pretty small. However, for the larger input depths,
since the amount of computation after distribution is more balanced,
we see a speedup. We discuss more in Section~\ref{sec:dist} about
this. Furthermore, we see that in most cases, spatial splitting
performs better. This is because spatial splitting, contrary to other
methods, does not have a significant communication overhead and since we only distribute on three devices, the number of
overlapping elements (i.e., additional computation in spatial splitting) are not high. This is why for larger $9\x9$ kernels, since the number of overlapping elements increases, the advantage of spatial splitting compared to other methods is less noticeable.



\section{Finding a Near-Optimal Distribution}

\label{sec:dist}
\begin{figure}[h]
  \vspace{-0pt}
  \includegraphics[width=1.0\linewidth]{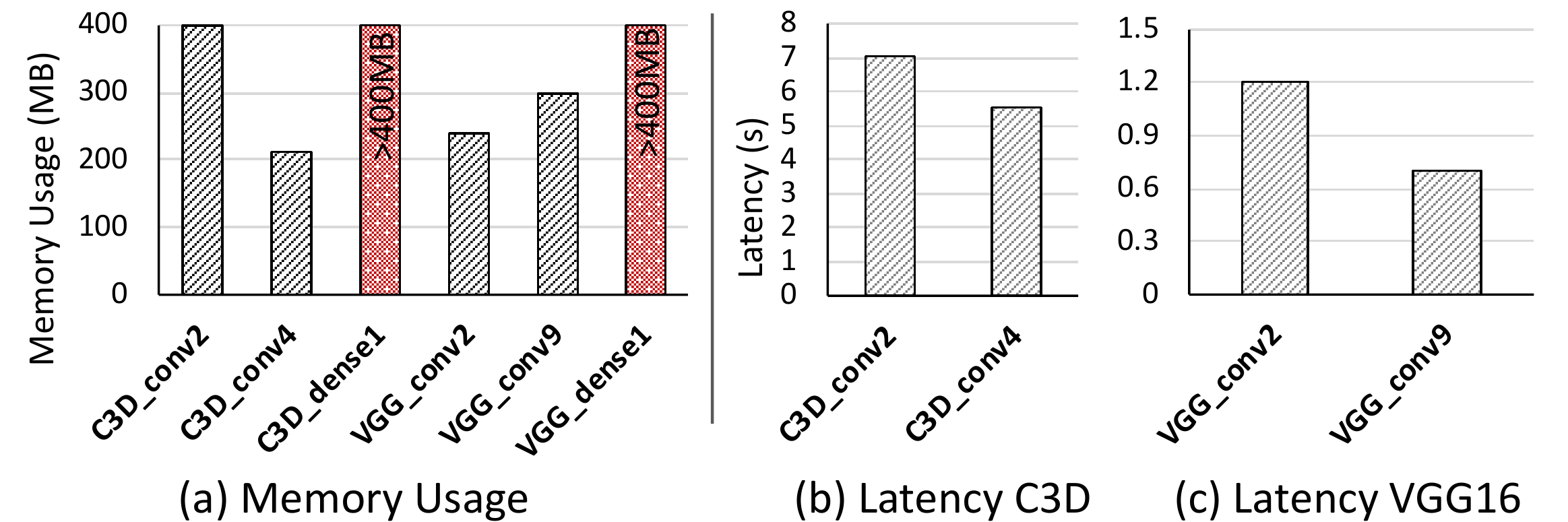}
  \vspace{-20pt}
  \caption{Memory usage and latency of VGG16 and C3D layers.}
  \vspace{-5pt}
  \label{fig:single-layer}
\end{figure}

 To understand why distributing and parallelizing DNN computations are necessary for resource-constrained devices Figure~\ref{fig:single-layer} shows memory usage and time to process an input (i.e., latency) of some layers in C3D and VGG16. As we see, beginning dense layers of both models have extremely long latencies (in order of minutes, not shown), because of their high memory footprint and low compute intensity which causes the usage of swap space. Hence, model parallelism is necessary for them. Convolution layers have much less memory footprint, but with a few layers on a device we will eventually exceed the available memory of the device and face the same issue as dense layers. Moreover, for convolution layers, the latency of a single computation is long and not suitable for real-time processing, as shown in Figure~\ref{fig:single-layer}b and c. Note that most DNN models have more than ten layer, and here we are only showing the statistics for one of them. The mentioned challenges are more exacerbated with more layers. In summary, total latency of executing the entire model on a single resource-constrained device is much longer because: (\romannum{1}) limited memory causes extremely slow swap space activities, and (\romannum{2}) latencies of all layers is accumulated because there are no parallelization opportunity. Model parallelism methods in the previous section helps us in solving these challenges because they reduce the memory footprint and exploit more compute resources.

Now that we are able to distribute and parallelize DNN computations using model and data parallelism, the question is: How can we find a near-optimal distribution for a given number of nodes? The distributed system that we study is essentially a  processing pipeline for DNN model. Our goal is to increase the execution performance of DNN models when performing single-batch inference in the terms of the number of inferences per second (IPS) (higher is better) and latency (lower is better). In general, if we have $\mathcal{W}$ amount of work and $n$ workers,  our speedup would be
\begin{equation}
  \footnotesize
  \text{Speedup} = \frac{\mathcal{W}}{\nicefrac{\mathcal{W}}{n} + \text{overhead}},
  \label{equ:speedup}
\end{equation}
in which the overhead entails to communication overhead ($\propto$ data size), and some fixed overhead such as the network set-up between devices. If the communication overhead dominates our distribution, then we will have a slowdown (as seen in the convolution layers study in Section~\ref{sec:layers}. To avoid such scenarios, we need to :(\romannum{1}) avoid unnecessary splits  to reduce the amount of communication overhead, and (\romannum{2}) associate enough work per node so the benefit of parallelizing exceeds communication overhead. To do so, we merge less compute intensive layers together on a single node.  We also monitor idle nodes and combine the layers, we also increase the utilization of each node thereby achieving a balanced pipeline. 


\textbf{Generating a Balanced Pipeline:}
To do a near-optimal distribution in our pipeline of IoT nodes, each node's latency should be similar to other nodes. Thus, the amount of work per node, or $\nicefrac{\mathcal{W}}{n}$ should be the same. Model parallelism helps us gain access to smaller granularities of work during distribution, therefore shorter latencies. On the other hand, data parallelism does not changes the amount of work per node, but increases the throughput. In other words, since the throughput increases (depending on the number of the nodes), the work on these data-parallelism  nodes could have a higher latency compared to other nodes in the pipeline. By considering these, to generate a distribution, first we need to create a database with a mix of (i) regression models based on the amount of work and type of the layers, and (ii) profiled data from some layers and their splitted versions (as seen in Section~\ref{sec:layers}). Then, we study our given DNN model layer by layer. If the memory footprint is large and causes swap activities, for that layer, we have to first use model parallelism. After that, we try to group less compute-intensive (sequential) layers to reduce the communication overhead mentioned before. The grouping is done in a way that the average latency for processing an input on each device would be similar. After deploying such initial distribution, we monitor the queue occupancy and latency of each device. With these gathered new data, we repeat the above steps and fine tune the distribution. Procedure~\ref{algo:algo} summarizes these steps.

\begin{center}
\begin{minipage}{1.0\linewidth}
\vspace{-3pt}
\begin{algorithm}[H]
   \caption{Heuristics for distributing a DNN model.}
   \label{algo:algo}
    \begin{algorithmic}[]
      \Procedure{GenerateDistribution}{} 
      \begin{flushleft}
          \textbf{Inputs:} list of layers $\varmathbb{L}$, \#Nodes $n$ \\
          \hspace{28pt} $mem_{size}$: Memory size per node \\
          \hspace{28pt} Regression models or profiling database, $\varmathbb{D}$ \\
          \textbf{Outputs:} dictionary of node IDs to a set of its tasks, $\varmathbb{T}$   
      \end{flushleft} 
      \vspace{-5pt}
      \EndProcedure
        \State \textbf{Step1:} Check memory usage all layers in $\varmathbb{L}$ using $\varmathbb{D}$, if larger than $mem_{size}$, add that layer to the model parallelism list.
        \State \textbf{Step2:} Using latency of layers in $\varmathbb{D}$ and their splitted version, and by ensuring sequential dependency of layers, try to create groups of layers with same latency. Create $\varmathbb{T}$.       
        \State \textbf{Step3:} Deploy $\varmathbb{T}$. Monitor queue occupancy and latency on each device. Goto Step2.

\end{algorithmic}
\end{algorithm}
\vspace{-5pt}
\end{minipage}
\end{center}

\section{System Evaluation}
 
 \label{sec:system}

We evaluate our method on a distributed system with Raspberry Pi 3s~\cite{pi3} (Table~\ref{tab:pi}). To show how our distribution heuristics provides better performance, we compare our results with a randomly assigned distributed system. The randomly assigned distributed system ensures correctness, but not an optimally designs a processing pipeline. In details, random assignment can create an unbalanced amount of work among the nodes, or assign a memory-intensive computation to one layer. For all implementations, we use Keras 2.1~\cite{chollet2015keras} with the TensorFlow backend (version 1.5)~\cite{tensorflow2015-whitepaper}. For RPC calls and serialization we use Apache Avro~\cite{apache}. A local network with the measured bandwidth of 94.1\,Mbps and a measured client-to-client latency of 0.4\,ms for 64\,B is used. All trained weights are loaded to each Pi's storage, so each Pi can be assigned to any task. 
\renewcommand{\arraystretch}{0.7}
\begin{table}[h]
    \small
	\centering
	\vspace{-10pt}
	\caption{Raspberry Pi 3 specifications.}
	\vspace{5  pt}
	\begin{tabular}{c | c | c}
		\toprule
        CPU & \multicolumn{2}{c}{1.2\,GHz Quad Core ARM Cortex-A53} \\
        Memory &  \multicolumn{2}{c}{900\,MHz 1\,GB RAM LPDDR2} \\
        GPU & \multicolumn{2}{c}{No GPGPU Capability} \\
        Price & \multicolumn{2}{c}{\$35 (Board) + \$5 (SD Card)} \\
		\bottomrule
	\end{tabular}
	\label{tab:pi}
	\vspace{-5pt}
\end{table} \renewcommand{\arraystretch}{1}
%
%
%
%

\begin{figure*}
  \vspace{-0pt}
  \includegraphics[width=1.0\linewidth]{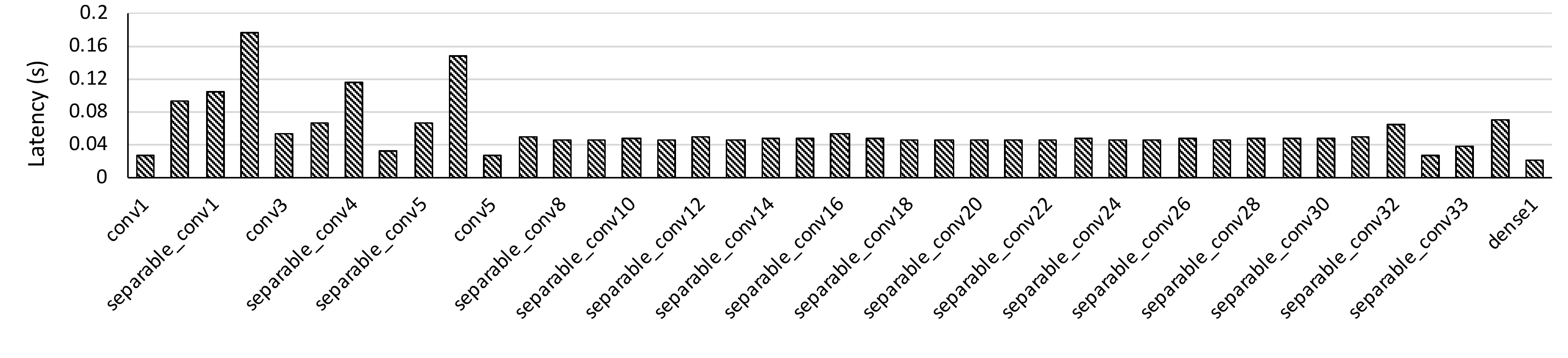}
  \vspace{-30pt}
  \caption{Xception layer-wise latency on Raspberry Pi.}
  \vspace{-5pt}
  \label{fig:xception-latency}
\end{figure*}

To create the pipeline, after finding a distribution of computations, we create a single file containing a dictionary of the IP addresses and their assigned computation. We upload the file to all nodes, and each node, by reading model description and its assigned computation, finds its position in the pipeline. After handshaking, which takes less than one minute, the system is ready for processing. During runtime, each node reports its latency and the histogram of request queue occupancy. By collecting such stats, we are able to find bottleneck nodes in our pipeline and create a more balanced pipeline, as Procedure~\ref{algo:algo} describes.

In the reminder of this section, we will analyze the application of the described model-parallelism methods on several models. Note that, if a model, after analyzing layer-by-layer requirement, does not require application of model-parallelism, we will not describe it here. This is because, data parallelism can help these models tremendously by grouping the layers together and copy the work on several nodes. For instance, Resnet50 contains several low-latency layers. Therefore, to distribute it, we can easily create groups of layers with even latencies and use data parallelism to increase its performance.

\begin{figure}[h]
  \vspace{-0pt}
  \includegraphics[width=1.0\linewidth]{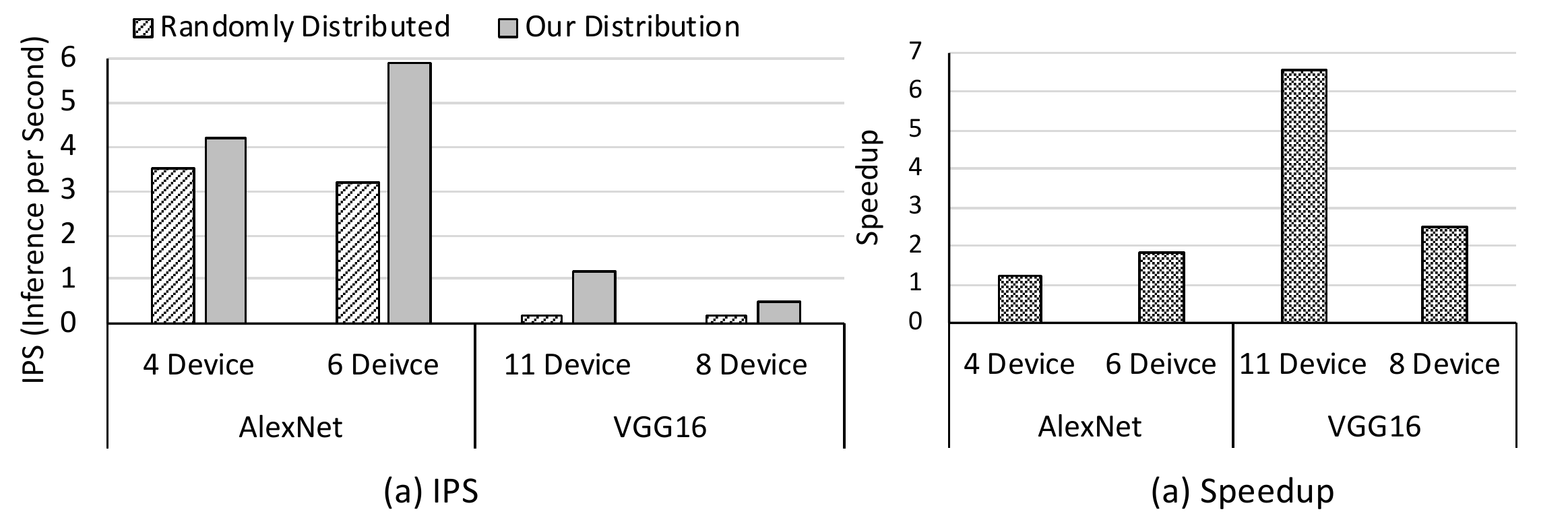}
  \vspace{-22pt}
  \caption{AlexNet and VGG16 result on a distributed system.}
  \vspace{-2pt}
  \label{fig:alexnet-vgg-with-fc}
\end{figure}

\begin{figure}[h]
  \vspace{-0pt}
  \includegraphics[width=1.0\linewidth]{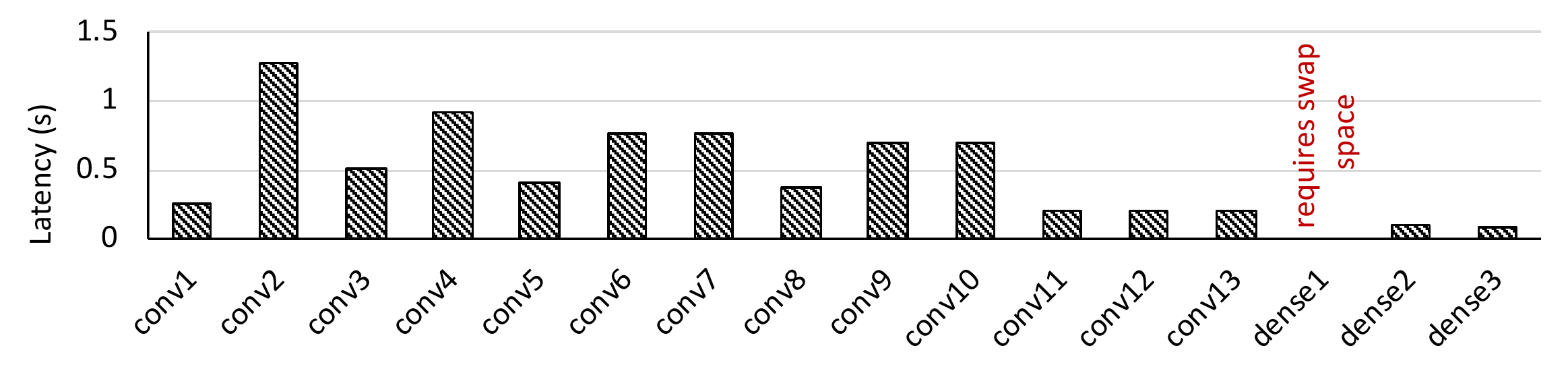}
  \vspace{-25pt}
  \caption{VGG16 layer-wise latency on Raspberry Pi.}
  \vspace{0pt}
  \label{fig:vgg-layer-latency}
\end{figure}

\textbf{AlexNet \& VGG16:} In these set of experiments, we deploy the entire AlexNet and VGG16, including last dense layers,  models on distributed systems. AlexNet, as introduced in Section~\ref{sec:background}, has 8 layers in total. Since the first dense layer in AlexNet face limited memory issue, all of our distributions perform output splitting for this layer. The rest of convolution layers are allocated to idle nodes. Our two example systems has 4 and 6 devices and achieve around 2${\times}$ speedups compared to randomly distributed systems, as seen in Figure~\ref{fig:alexnet-vgg-with-fc}. Because AlexNet layers all have low compute requirements (per layer latency of less than 0.2s), we could not get more benefit by distributing the computations. But, VGG16, compared to AlexNet, consists of more computationally intensive layers. Therefore, as Figure~\ref{fig:alexnet-vgg-with-fc} shows, we use 8 and 11 devices for distribution to achieve up to 6${\times}$ speedup. Note that, similar to AlexNet, since we include the first dense layer, all of our distributions perform output splitting for this layer. For other layers, to gain a better insight, in Figure~\ref{fig:vgg-layer-latency}, we measured layer-wise latency of VGG16 layers that are executed on Raspberry Pi. Except the first dense layers, we are able to run all other layers on a single Raspberry Pi. But, some layers have extremely long latency, so we will be bounded by such layers in our pipeline (e.g., second convolution layer). Our 11- and 8-device systems bypass this bottleneck by using the methods proposed in Section~\ref{sec:layers} to achieve these speedups.

\begin{figure}[h]
  \vspace{-0pt}
  \includegraphics[width=1.0\linewidth]{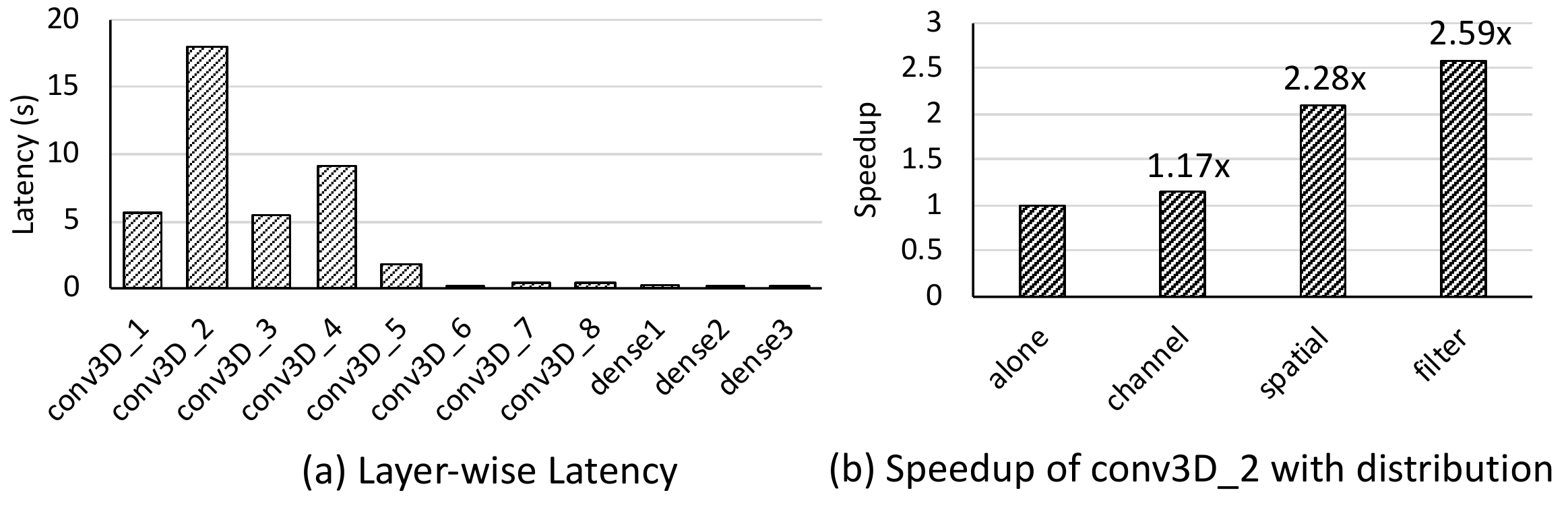}
  \vspace{-20pt}
  \caption{C3D layer-wise latency and applying model-parallelism methods on the heaviest layer.}
  \vspace{-2pt}
  \label{fig:c3d-layer-latency}
\end{figure}

\textbf{C3D:}
To understand when applying model-parallelism methods are appropriate, we analyze layer-by-layer latency of C3D models on the Raspberry Pi in Figure~\ref{fig:c3d-layer-latency}a. As shown, C3D is quite heavy for resource-constrained devices which is due to its layers that have high latency. This high latency is caused by a few convolution layers. The memory footprint of these convolution layers can be fit in our device memory, thus, the latency is caused by their heavy computations. In other words, because C3D has 3D convolutions, a key layer in understating temporal content, which has high computation demands, our model will experience a huge slowdown when begin distributed. To see how model parallelism can help these situations, we apply our three methods of model-parallelism with three devices on the second convolution layer. As seen, we can get up to 2.6${\times}$ speedup by using only three devices.  

\textbf{Resnet50 \& Xception:}
We did the similar latency analysis for Resnet50 model. However, as seen from their model in Section~\ref{sec:background}, since the layer-wise latency of each layer is so short (less than 0.2 seconds per layer), there is not much opportunity in applying model parallelism on these model. Figure~\ref{fig:xception-latency} provides a detailed latency overview of Xception. As seen the highest latency is less than 0.2 seconds. In this case, data-level parallelism will provide a linear speedup as the number of nodes. 
%
%
%
%

\section{Conclusion}

\label{sec:conclusion}

In this work, we proposed a solution to aid moving the computations of DNNs closer to the edge devices. Our target was resource-constrained devices such as prevalent IoT that have small memory and low computation power. We increased the real-time performance of single-batch inferencing by deploying a processing pipeline that exploits the collaboration between such devices. To overcome the limited memory and compute power of these devices, we introduce several model-parallelism methods and throughly analyzed their cost and benefits. Finally, we deployed processing pipelines for a few state-of-the-art visual models. for For future work, we plan to extend our work to more than visual DNNs, covering areas such as translation and speech recognition. Furthermore, we are studying the possibility of various methods in alleviation the communication overhead such as bypassing merging in the next layer, compression, and using coded distribution. 


\newpage
\bibliography{ref}
\bibliographystyle{sysml2019}


\end{document}